\pdfoutput=1

\documentclass[11pt]{article}

\usepackage[final]{acl}

\usepackage{times}
\usepackage{latexsym}
\usepackage{soul}
\usepackage{xcolor}
\usepackage{float}
\usepackage{subcaption}
\usepackage{listings}
\usepackage{changepage}
\usepackage{tcolorbox}

\usepackage[T1]{fontenc}

\usepackage[utf8]{inputenc}

\usepackage{microtype}
\usepackage{amsmath}
\usepackage{inconsolata}

\usepackage{graphicx}
\usepackage{xcolor}

\title{From RAGs to rich parameters: Probing how language models utilize external knowledge over parametric information for factual queries}

\author{
  \textbf{Hitesh Wadhwa\textsuperscript{1, *}},
 \textbf{Rahul Seetharaman\textsuperscript{1, *}},
  \textbf{Somyaa Aggarwal\textsuperscript{1, *}},
  \textbf{Reshmi Ghosh\textsuperscript{2}},\\
  \textbf{Samyadeep Basu\textsuperscript{2, 3}}, 
\textbf{Soundararajan Srinivasan\textsuperscript{2}},
  \textbf{Wenlong Zhao\textsuperscript{1}},
  \textbf{Shreyas Chaudhari\textsuperscript{1}},
  \\
  \textbf{Ehsan Aghazadeh\textsuperscript{1}}
\\
 \textsuperscript{1}University of Massachusetts, Amherst,
 \textsuperscript{2}Microsoft,
\textsuperscript{3}University of Maryland, College Park
\\
\textsuperscript{*}\small{Equal Contributions} ||
\small{
  \textbf{Correspondence:} \href{mailto:reshmighosh@microsoft.com}{reshmighosh@microsoft.com}
 }
}

\begin{document}
\maketitle
\begin{abstract}
Retrieval Augmented Generation (RAG) enriches the ability of language models to reason using external context to augment responses for a given user prompt. This approach has risen in popularity due to practical applications in various applications of language models in search, question/answering, and chat-bots. However, the exact nature of how this approach works isn't clearly understood. In this paper, we \textbf{\textit{mechanistically}} examine the RAG pipeline to highlight that language models take  ``shortcut'' and have a strong bias towards utilizing only the context information to answer the question, while relying minimally on their parametric memory. We probe this mechanistic behavior in language models with: (i) Causal Mediation Analysis to show that the parametric memory is minimally utilized when answering a question and (ii) Attention Contributions and Knockouts to show that the last token residual stream do not get enriched from the subject token in the question, but gets enriched from other informative tokens in the context. We find this pronounced ``shortcut'' behaviour true across both LLaMa and Phi family of models.
\end{abstract}

\section{Introduction}



With the burgeoning use of Language Models (LMs) in many industrial applications, retrieval Augmented Generation (RAG) has become popular as a mechanism of providing additional context for effective \textit{reasoning} to mitigate \textit{hallucinations}. Yet, the usefulness of RAG to provide meaningful information in comparison to model priors is an under-explored area of research. On the other hand, knowledge localization and editing techniques\cite{wang2024editing}\cite{wang2024detoxifying}\cite{gupta2024unified}\cite{gupta2024model}\cite{sharma2024locating}\cite{conmy2023towards}\cite{wu2024interpretability} in LMs such as ROME \cite{meng2022locating} and MEMIT \cite{meng2022mass} are traditionally focused on adjusting the internal parameters of the LMs to update or correct knowledge. However, a mechanistic understanding of how RAG context influences LM predictions over prior knowledge hasn't been studied till date. And the rise of RAG usage necessitates us to understand quantitatively the interplay between the LM's prior knowledge and the external information retrieved during inference, for preventing drift in model reasoning.

\begin{figure}[!h]
    \centering\includegraphics[width = 0.5\textwidth]{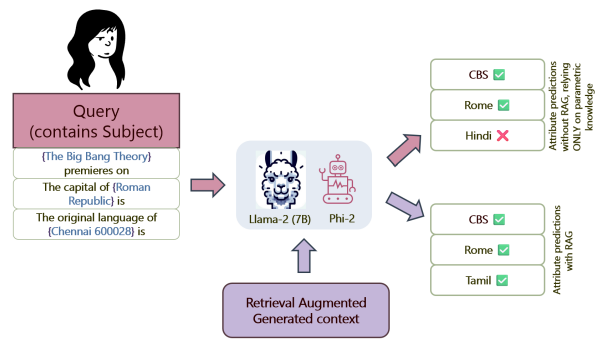}
    \caption{\label{process} \small
    Setup of a factual QA system with RAG, utilized in this paper, for understanding the usefulness of parameteric knowledge stored in LlaMa and Phi.}
\end{figure}
In this paper, we aim to \textbf{analyze} and \textbf{interpret} the dependency of LMs  on parametric knowledge versus the  retrieved information presented via RAG. Towards this goal, we rely on established methods of locating factual knowledge stored in the model parameters.   

We find that: (i). Parametric knowledge is minimally used within Multi Layer Perceptrons(MLPs) in the presence of retrieved context. and (ii). The last token residual stream, crucial for forming the final output, derives more enriched information from the attribute token present explicitly in the context rather than from the subject token within the query. These insights highlight a pronounced "shortcut" behavior in LMs, where the models prioritize external context over internal knowledge. Through this analysis, our work contributes to the a novel understanding of the mechanisms underlying LMs' preference for the information provided via RAG. 



\section{Related Work}
RAG systems \cite{lewis2021retrievalaugmented} have become popular in practical natural language systems as they significantly improve the performance of LM applications by integrating external context\cite{shao2023eragent}\cite{singh2023domain}\cite{ingestai2023rag}\cite{kaddour2023challenges}
\cite{chen2024benchmarking}\cite{ren2023investigating} 
However, utilizing RAGs can also have nuanced outcomes such as generation of inconsistent predictions, even with perfect retrieval results\cite{Hagstrm2023TheEO}.\cite{wu2024faithful} explore the role of RAG in reducing hallucinations and enhancing accuracy in large language models such as GPT-4, building on prior work\cite{lewis2021retrievalaugmented}\cite{shuster2021retrieval} that leverage external retrieval systems to mitigate model errors. Even though RAG models are extensively used, and their shortcomings documented, only~\cite{wu2024faithful} delves into the balance between a model’s internal knowledge and externally retrieved information, examining their practical value. However, a systematic \textbf{mechanistic} exploration of model's preference for RAG-provided information over their parametric knowledge contribution has not yet been conducted, to the best of our knowledge. Our study mechanistically probes into the internal workings of large language models and how they exhibit a "shortcut mechanism" when they are provided with non-parametric knowledge via a RAG system.

\section{Probing Mechanisms}
\vspace{-3mm}
To mechanistically interpret the knowledge contributions towards factual reasoning by LLMs and SLMs, we use three methods for causal mediation, described as follows:
\subsection{Causal Tracing}
Causal tracing \cite{meng2022locating}identifies specific hidden states that significantly influence factual predictions. The approach involves three steps - a clean run, corrupted run and a corrupted-with-restoration run. The corrupted run involves corrupting a certain span of the text, and running the forward pass of the model. In the restoration run, activations from the clean run are patched one by one into the corrupted run, and the increase in answer probability is observed; the most crucial activations are thus causally determined.

Finally, the \textbf{Indirect Effect (IE)} of a specific hidden state \( h^{(l)}_i \) is defined as the difference between the corrupted run and the corrupted-with-restoration run probabilities: $
\text{IE}(h^{(l)}_i) = P^*_{\text{clean}}(h^{(l)}_i)[y] - P^*[y]$
and by averaging these effects over a sample, the \textbf{Average Indirect Effect (AIE)} is computed for all hidden states, providing a quantitative measure of their importance in factual prediction.

\subsection{Attention Knockout and Contribution Mechanism}
\vspace{-2mm}

The \textbf{Attention Contribution} \cite{yuksekgonul2024attention}, focuses on the role of attention mechanisms in shaping the output of language models. This approach investigates how attention weights, particularly from the subject token in a query to the last token position, contribute to the model's predictions. By examining the norm of these attention weights \( \|a^{(\ell)}_{i,T}\| \), we observe what tokens the last token pays the most attention to, during the generation process. See appendix \ref{sec:attention-contribution} for norm calculation details.
The \textbf{Attention Knockout} mechanism \cite{geva2023dissecting} identifies critical attention edges in transformer-based models that are essential for maintaining prediction quality. The process involves identifying critical edges whose removal significantly degrades the model's prediction quality. To test the importance of these edges, attention weights between two positions $r$ and $c$ at a layer $l$ are set to negative infinity:  $M^{l+1,j}_{rc} = -\infty \quad \forall j \in [1, H]$


\begin{figure*}[t!]
    \centering
    \begin{subfigure}[b]{0.49\textwidth}
        \centering
        \includegraphics[width=\textwidth]{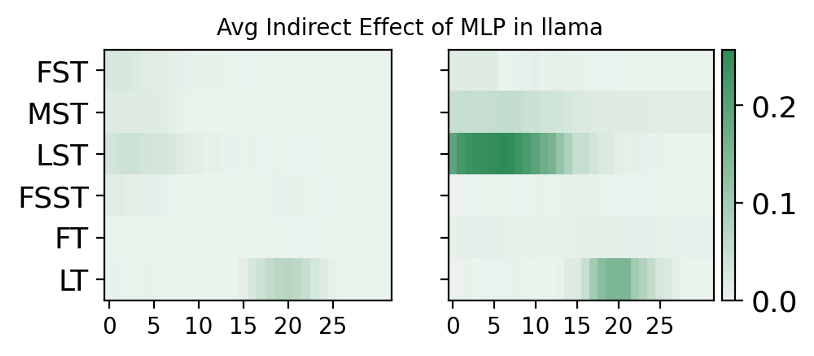}
        \caption{LLaMa-2 with RAG vs. LLaMa-2 Vanilla}
        \label{fig:first_image}
    \end{subfigure}
    \hfill 
    \begin{subfigure}[b]{0.49\textwidth}
        \centering
        \includegraphics[width=\textwidth]{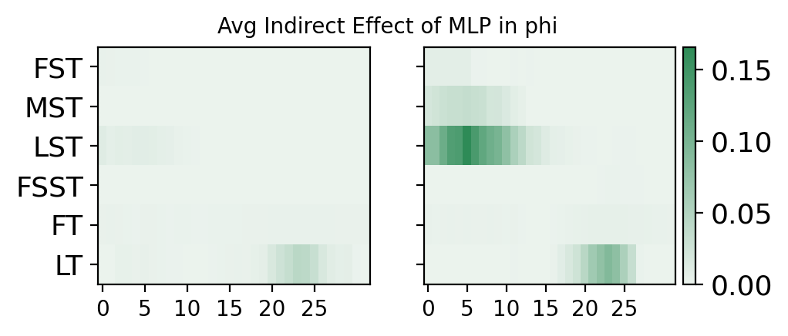}
        \caption{Phi-2 with RAG vs. Phi-2 Vanilla}
        \label{fig:fourth_image}
    \end{subfigure}
    \caption{\textbf{Language models minimally rely on the MLP parametric memory in the presence of retrieved context}. From left to right: Average Indirect Effect from MLPs after corrupting subject + context for scenario based on RAG and subject for vanilla-case. Here, FST=\textbf{\textit{First Subject Token}}, MST=\textbf{\textit{Middle Subject Tokens}}, LST= \textbf{\textit{Last Subject Token}}, FSST=\textbf{\textit{First Subsequent Token}}, FT= \textbf{\textit{Further Tokens}}, LT= \textbf{\textit{Last Token}}. On average 5 times decrease in AIE is observed for LST with RAG vs. vanilla, signalling decrease in usage of MLP when RAG context present.}
    \label{fig:final_causal_trace_plots}
\end{figure*} This prevents the source position $x^l_r$ from attending to the target position $x^l_c$, blocking information flow at that layer. The degradation in prediction quality after blocking attention edges identifies which edges are critical for information flow.



\section{Datasets and Models}
\vspace{-2mm}
\subsection{Models}
\vspace{-1.8mm}
For a comprehensive mechanistic probing, we leverage two state-of-the-art LMs, Phi-2 (2.7B) \cite{textbooks2} and LLaMA-2 (7B) \cite{touvron2023llama} models, which were trained on different corpora. Difference in parametreic knowledge between two different family of models, allows us to comprehensively probe the influence of RAG for factual queries in scenarios involving these models. Also chosing open-source LMs enables us measure causal mediation easily.

\subsection{{Dataset}}
\vspace{-1.8mm}
In this paper, we scope the analysis to determine the influence of external information provided by RAG context against model priors, to only factual query predictions from aforementioned LMs. Thus, we utilize the \textbf{Knowns Fact Dataset} of 1209 factual queries, introduced in \cite{meng2022locating}. Each record in the dataset is of $(s,r,o)$ format of subject, relation and object/attribute, respectively \footnote{ subject  part of the user query. For example, for user query: "The Space Needle is located in the city of" the subject will be defined as "The Space Needle". When we say attribute or object, we mean the answer to that query which will be present only once in the context generated placed at the first segment. Example can be found in Appendix \ref{sec:appendix-syn-prompts}.}.

For the RAG dataset, we synthetically generate RAG context for each query from the Knowns-Fact dataset using GPT4. This was done to control the variables such as length of each segment within the RAG context and the presence of \textbf{\textit{attribute}} or \textbf{\textit{object}}. Further details on prompts used and samples from dataset in Appendix \ref{sec:appendix-prompts}. 
In the scope of this work, we work with a vanilla setting, where no RAG context is present for queries to get enriches, and a RAG setting.
The generation was made sure to follow our constraints using quality assurance techniques which regenerated the context when the constraints were not satisfied. The code can be found here in Appendix \ref{sec:QA code}


\section{Empirical Results}
\vspace{-2mm}
\subsection{Finding 1: Language models minimally use parametric memory in the presence of context}
We start by mechanistically probing the contributions of various MLP layers for Llama-2 (7B) and Phi-2 for a representative set of randomly sampled prompts\footnote{We randomly select a small subset, 50 prompts as causal tracing with RAG context takes significant time to experiment with, in the order of a 4-5 hours for 20 word segments of 5 count} for both scenarios, i.e, vanilla vs. RAG to understand fact prediction. For RAG scenarios, the entire \textbf{\textit{context}} along with \textbf{\textit{subject}} is corrupted as part of causal tracing, whereas for the vanilla case only \textbf{\textit{subject}} is corrupted. 
Figure \ref{fig:final_causal_trace_plots} presents the decrease in AIE in presence of RAG of the LST as compared to vanilla(no RAG) setting.

\begin{figure*}[!h]
    \hspace{-0.5cm}
    \includegraphics[width = 1.045\textwidth]{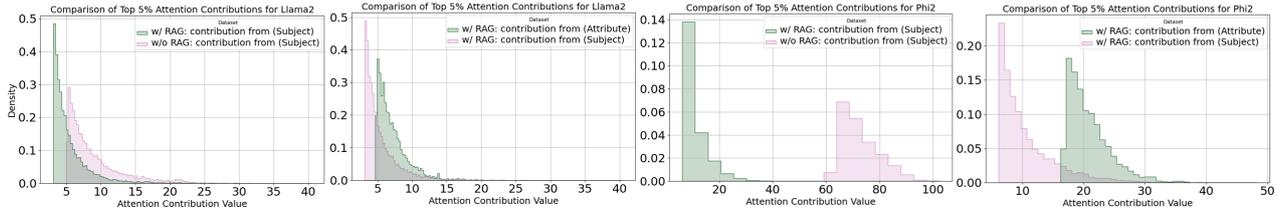}
    \caption{\textbf{The last token residual stream obtains less enriched information from the subject token in the query in the presence of retrieved context.}\small{(a) Subject Token contribution for RAG vs vanilla in Llama-2, (b) Comparison of subject and attribute contributions w/ RAG for Llama-2, (c) Subject contribution for RAG vs vanilla in Phi-2, (d) Comparison of subject and attribute contributions w/ RAG for Phi-2. 4a. and 4c indicates subject contribution is twice as lower in case of RAG as compared to vanilla. 4b and 4d shows that attribute token's attention contribution is 5 times more than the subject contribution.}} 
    \label{fig:finalAttentionContributions}
\end{figure*}

 We analyze  the Average Indirect Effect of MLPs representing subject tokens and compare against vanilla vs. RAG context scenarios for Llama-2(7B) for 50 examples from the knowns fact dataset, and find that the AIE decreases 5 times (from ~0.2 to ~0.0375), proving that subject tokens within the query does not elicit the parametric memory when the context is present. 
Similarly, for the case of a smaller language model such as Phi-2, we have a similar observation where we find that the language model does not use the parametric memory. This is in contrast to a non-RAG, vanilla case where the subject token has a high AIE and serves as a hotspot of factual retrieval from parametric memory. In addition to the MLPs, we also perform causal tracing on attention layers, details of which can be found in Appendix \ref{sec:causaltracing}

\subsection{Finding 2: Last token residual stream obtains more enriched information from the context, rather than subject token in query}

Inspired by findings of a strong attention contribution from the Subject Token (ST) in the query question to the Last Token (LT) position for factual queries in\citep{yuksekgonul2024attention}, we try to uncover  
any signal of relevant information transfer between subject token and the last token position in LMs for factual queries.We compute the Attention Contributions from ST \footnote{ST refers to the subject tokens of the user query.} to the LT for LlaMa-2 and Phi-2 for vanilla and RAG scenarios 
for all 1209 factual queries in Knowns Fact Dataset. We find that 70\%  of the layers don't contribute to the final token prediction and therefore resulting in almost 0 contribution to the Last Token (LT). Thereby, as shown in Figure \ref{fig:finalAttentionContributions} we extract the top 5$\%$ of the Attention Contributions from the ST to the LT for vanilla vs. RAG scenarios using LlaMA and Phi to amplify the difference. We observe that Specifically for Fig\ref{fig:finalAttentionContributions}.a and Fig \ref{fig:finalAttentionContributions}.b, the Attention Contributions from Subject Token decrease in the presence of RAG indicating, the larger influence of RAG context in predicting facts.For LLaMa-2, the mean attention contribution for RAG case is 5.6094 vs. 9.0054 in vanilla setting. For Phi, Attention Contribution at ST is 10.6650 for RAG vs. 72.5961 in the vanilla case, which 7 times larger.
\begin{figure}[!h]
    \hspace{-0.5cm}
    \includegraphics[width = 0.5\textwidth]{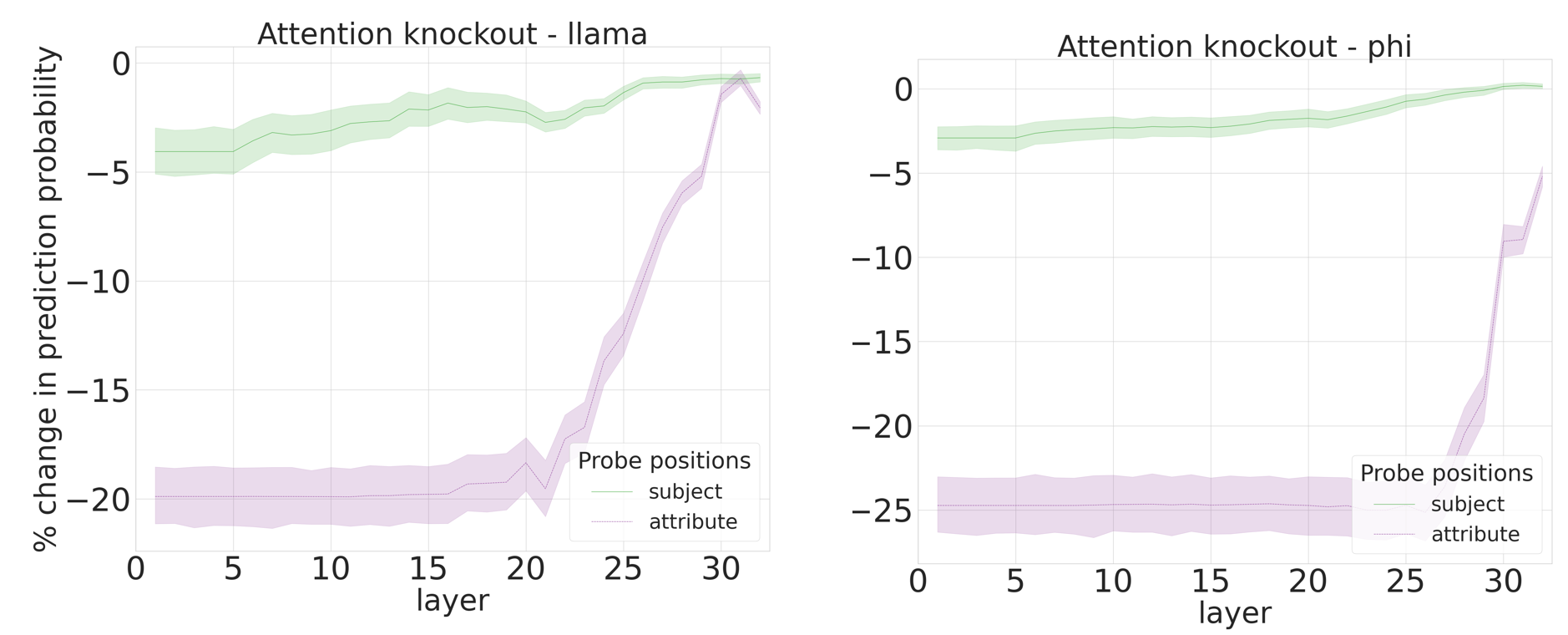}
    \caption{\label{process} \small
    \textbf{In the presence of retrieved context, knocking out attention weights from the subject in query to the last token has minimal effect}. (Left) Llama2 (Right) Phi2. [Knocking out attribute tokens decreases probability upto 25\%in Phi2 and 20\% in Llama2 and only 5\% probability is reduced on knocking out subject token attention.] }
    \label{fig:finalAttentionKnockouts}
\end{figure}

Additionally, we also analyze Attention Contributions for Attribute Tokens (AT)\footnote{Attribute tokens refers to the expected answer of the query being asked, present in the RAG context, which is also the same as the \textbf{object}}, and compare them against ST. The controlled RAG context we generated synthetically ensures there is \textbf{only one} AT present in the context. 
We find in Fig \ref{fig:finalAttentionContributions}.b, and \ref{fig:finalAttentionContributions}.d, when compared against Attention Contributions of AT present in RAG context, against ST in the query, AT has a larger influence in fact predictions. For LlaMa-2, the mean attention contribution at AT is 7.1242, while at ST is 5.6094. For Phi-2, it is 20.8902 and 10.6650, respectively, i.e, 2 times higher than at ST.


To validate this finding further, we use Attention Knockouts \cite{geva2023dissecting} to measure the change in probability of the predicted token (object/attribute), when the attention weights from the ST in the query to the last token is knocked off. Figure \ref{fig:finalAttentionKnockouts} presents that for the RAG scenario, knocking off attention weights from the subject in query to the last token leads to a probability drop of less than 5 percent in both LLaMa-2 and Phi-2. However, we observe a much stronger drop in the probability of the original predicted token, (20\%)in LLaMa-2 and 25\% in Phi-2. These results highlight that in presence of RAG context, the last token residual stream ignores information from the subject token position in the query and instead solely relies on the token contributions from the context. Additionally, we perform knockouts in the vanilla setting on the subject token(details in Appendix \ref{sec:knockouts_vanilla}.)
\begin{tcolorbox}
\textbf{Main Takeaway}: In the presence of retrieved RAG context, language models internally rely primarily on the context, while minimally using the parametric memory to answer a question.
\end{tcolorbox}
    \label{fig:Attention_Knockouts}

\section{Discussion and Conclusions}
\vspace{-2mm}

This paper is the first study to utilize three different mechanistic probing methods to understand the benefits of using RAG context as an external knowledge source to complement the parametric knowledge stored in the models as prior for factual queries. Our work explores the utility of parametric memory, and the interplay between parametric and non-parametric memory in the process of retrieval augmented generation.We find that parametric memory becomes less critical for factual recall when RAG context is augmented to the prompt. Through attention contributions, attention knockouts and causal traces, we specifically observe a reduced reliance on the subject token, and the MLP activations associated with it, when the context is augmented with RAG.

\section{Limitations and Future Work}
Our study is limited by the analysis using short RAG-based context. Handling really long context currently incurs a prohibitively large computational overhead in causal tracing. We plan to study the impact of long context and the impact of subject token and attribute token with respect to position and the tendency to exhibit proximity and recency bias \cite{liu-etal:2023:arxiv} in a future work. In addition, similar analysis of instruction tuned models and models that are finetuned on objectives like RLHF is a topic for future work. The current study involves a well controlled setting where attribute token is present only once in the context and the context itself is synthetically generated and well-formed. Retrieved outputs, in practice is very noisy and often sensitive to the quality of the retrievers, rankers, and the hyperparameters used. Examining those is also a natural extension of this work.



\bibliography{acl_latex}

\appendix

\section{Sample Data from Known Facts Dataset}
\label{sec:appendix-prompts}




\begin{verbatim}
{
"known_id": 14,
"subject": "Eavan Boland",
"attribute": "Dublin",
"template": "{} was born in",
"prediction": " Dublin, Ireland, in 1971.
He is the",
"prompt": "Eavan Boland was born in",
"relation_id": "P19"
}
\end{verbatim}


\section{Sample Data from synthetically generated GPT4 Dataset with RAG contexts}
\label{sec:appendix-syn-prompts}


\begin{verbatim}
{"index": 14, 
"user_query": "Eavan Boland was born in", 
"object": "Dublin", 
"response": ["Boland was born in Dublin, 
Ireland, 1944, and became a leading voice 
in contemporary Irish poetry, 
exploring women's",
"Her birthplace greatly influenced her 
works, emphasizing historical narratives 
and the role of women in Irish society
through poetry.",
"Boland's early life in Ireland shaped her
poetic voice, focusing on national 
identity, gender issues, and 
personal history.",
"Educated at Trinity College, her 
surroundings nurtured her literary 
genius, leading to a profound
impact on modern literature.",
"Despite her global travels and 
international teaching positions, her 
Irish roots remained central to 
her thematic concerns in poetry"]
}
\end{verbatim}

\textbf{Initial Query :}
\begin{tcolorbox}
    Eavan Boland was born in
\end{tcolorbox}
\textbf{Query Augmented with RAG context :}
    \begin{tcolorbox}
     Information is below:---------------- \\ Eavan Boland was born in Dublin, Ireland, 1944, and became a leading voice in contemporary Irish poetry, exploring women's\\ Her birthplace greatly influenced her works, emphasizing historical narratives and the role of women in Irish society through poetry.\\ Boland's early life in Ireland shaped her poetic voice, focusing on national identity, gender issues, and personal history.\\ Educated at Trinity College, her surroundings nurtured her literary genius, leading to a profound impact on modern literature.\\ Despite her global travels and international teaching positions, her Irish roots remained central to her thematic concerns in poetry.\\ Given the context information and not prior knowledge, complete the following: \\\\ Eavan Boland was born in
\end{tcolorbox}

\textbf{Prompt used for generation of synthetic dataset: }

\textcolor{blue}{System Prompt for GPT-4}
\begin{tcolorbox}
You are an expert data generation bot, specializing in generating 20 word segments.

- You generate these 20-word segments by consolidating information/knowledge AROUND a sentence that the user provides, that is: [\textcolor{blue}{user query}] [\textcolor{blue}{object}].

- While generating these five 20-word segments based on the sentence provided by the user, here: [\textcolor{blue}{user query}] [\textcolor{blue}{object}], make sure that only 1 of the 5 segments has the [\textcolor{blue}{object}] explicitly mentioned. FOLLOW THIS INSTRUCTION STRICTLY.

- Also make sure that none of these segments contain: [\textcolor{blue}{user query}]. Double check to make sure this instruction is strictly followed.

- Also make sure that these segments follow the format of an array of segments, i.e, [segment1, segment2, segment3, segment4, segment5]
\end{tcolorbox}

\textcolor{blue}{User Prompt for GPT-4}
\begin{tcolorbox}
    Generate five 20-word segments based on the following sentence: [\textcolor{blue}{user query}] [\textcolor{blue}{object}]
\end{tcolorbox}

The RAG-like dataset of augmented contexts is created synthetically by prompting GPT-4. We also experimented with an actual RAG pipeline, with documents from wikipedia along with the existing query set. However we observed that using a RAG pipeline comes with its own disadvantages with respect to controllability. Given the sensitity of the output measures like AIE, probabilities, etc to inputs and their perturbations, using a RAG pipeline adds more variability, as retrieved documents can be noisy and extremely sensitive to the underlying retrieval model and its hyperparameters.







\section{Background}
\label{sec:attention-contribution}
\subsection{Attention Contribution}
\cite{yuksekgonul2024attention} 
introduced SAT-Probe, 
to predict constraint satisfaction and factual errors 
by leveraging self-attention patterns to determine if generated text adheres to specified constraints and measuring 
the contribution of different components to the model's predictions. 


And \textbf{Attention to Constraints} is achieved by 1. identify constraint tokens within the input, 2. tracking the attention weights \( A^{(\ell)}_{i,j} \) ( \( \ell \) is layer, \( i \) is query token and \( j \) is constraint token), 3. aggregating attention weights across layers and heads to determine attention contribution \( A_{Ck,T} \) (where \( Ck \) is constraint tokens \& \( T \) is the entire token set).





Finally, the norm of attention contributions \( \|a^{(\ell)}_{i,T}\| \) from constraint tokens \( c \) to target token \( T \) at layer \( \ell \) is measured by aggregating  these norms across all layers and heads to form a comprehensive metric for attention contribution.
\[
a^{(\ell,h)}_{c,T} = A^{(\ell,h)}_{c,T}(x^{(\ell-1)}_c W^{(\ell,h)}_V)W^{(\ell,h)}_O
\]
where \( a^{(\ell,h)}_{c,T} \) indicates the attention contribution from a constraint token \( c \) through head \( h \) to the final token \( T \). The total contribution is:
\[
a^{(\ell)}_{c,T} = \sum_{h} a^{(\ell,h)}_{c,T}
\]
For multiple constraint tokens, the maximum value is considered:
\[
A^{(\ell,h)}_{C,T} = \max_{c \in C} A^{(\ell,h)}_{c,T} \quad \text{and} \quad a^{(\ell,h)}_{C,T} = \max_{c \in C} \|a^{(\ell,h)}_{c,T}\|
\]

\textbf{Correlation with Factual Correctness}

Analyze the correlation between the aggregated attention norms and the factual correctness of the model's outputs. Higher attention norms to constraint tokens are found to correlate with increased factual accuracy, providing a predictive measure for evaluating the reliability of the model's responses.

\section{Attention Knockouts}
\label{sec:knockouts_vanilla}
The attention knockouts \cite{geva2023dissecting} study the impact knocking out attention from a token position $i$ to $j$, where $i$ $\leq$ $j$ for an autoregressive model. More specifically, \cite{geva2023dissecting} study the impact of knocking out attention from the last token to the subject token, with prompts from the Knowns 1000 dataset, which is a dataset of queries in the form of $(s,r,o)$ triples. In addition to the attention knockouts in the RAG setting, we implement the attention knockouts on the subject token in the vanilla setting.

\begin{figure}[H]
    \centering
    \includegraphics[width=0.5\linewidth]{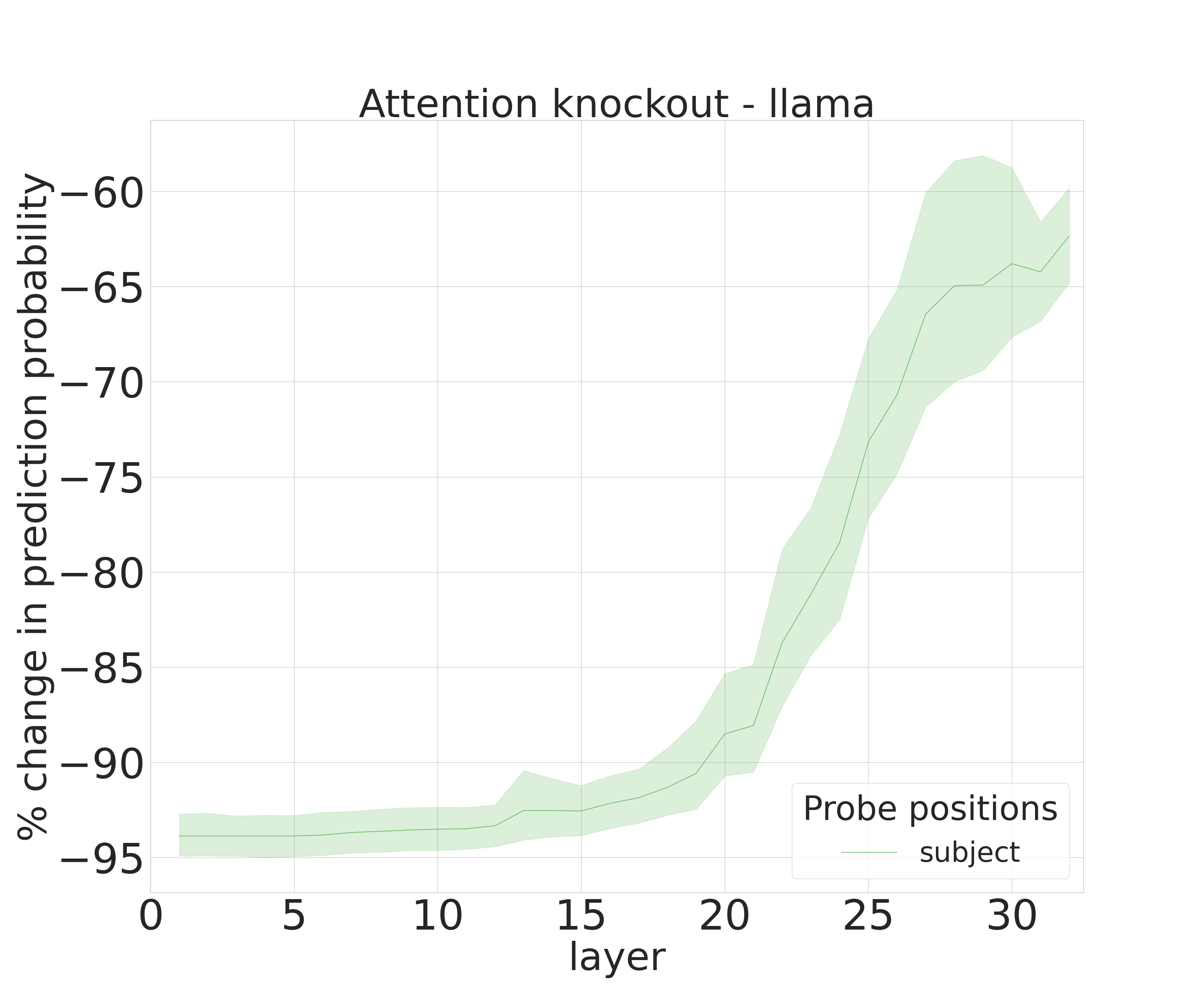}
    \caption{Attention knockouts in LLaMa - vanilla setting}
    \label{fig:llama_vanilla_knockout}
\end{figure}

\begin{figure}[H]
    \centering
    \includegraphics[width=0.5\linewidth]{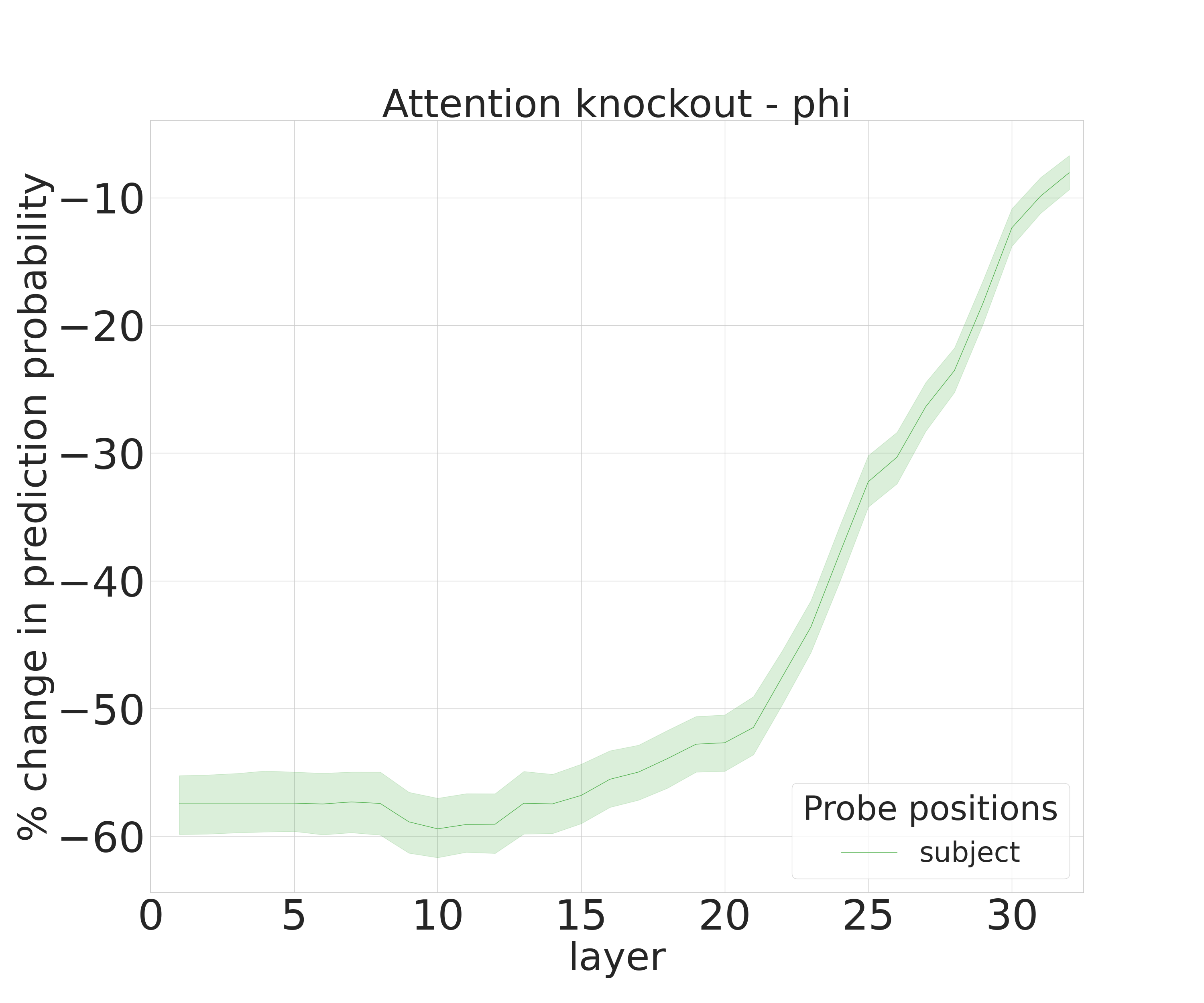}
    \caption{Attention knockouts in Phi - vanilla setting}
    \label{fig:phi_vanilla_knockout}
\end{figure}

Figure \ref{fig:llama_vanilla_knockout} and \ref{fig:phi_vanilla_knockout} show the attention knockout on the subject token in the vanilla setting. In the absence of added RAG context, we observe a 95 percent decrease in attribute probability in LLaMa and nearly a 60 percent decrease in the attribute probability in Phi-2. In the absence of external context, the model is reliant on parametric memory to answer the factual query, and hence the large probability drop on knocking out subject token attention.

\section{Quality checks on the generated synthetic data}
\label{sec:QA code}
Our data generation process comprises prompting GPT-4 to generate synthetic RAG context. The quality check primarily involves verifying the attribute token occurs exactly once within the generated context. The following piece of code is used to perform the verification.

\begin{lstlisting}[language=Python, basicstyle=\tiny]
    def isEntryOkay(entry):
        user_query = entry['user_query']
        object_value = entry['object']
        response = entry['response']
    
        # Check if object is present only once in the response
        object_count = response.count(object_value)
    
        # Check if user query is not present in the response
        query_in_response = user_query in response
    return object_count == 1 and not query_in_response

\end{lstlisting}

\section{Causal Tracing}
\label{sec:causaltracing}
The following positions are tracked while plotting the Average Indirect Effect (AIE). First subject token (FST), Middle Subject Token (MST), Last subject token (LST), Further Subsequent token (FSST), Further tokens (FT), Last token (LT). The last token is crucial to study, as it is projected onto a vocabulary during decoding. The last token residual is where information gets written during factual recall (both RAG and non-RAG). The last subject token positions are hotspots of parametric knowledge and factual recall in the vanilla non-RAG setting. Besides, due to causal attention, last subject token (LST) is equipped with context about First (FST) and Middle subject tokens (MST) as well.Further tokens (FT), Further Subsequent tokens (FSST) are not found to have significant causal impact in both RAG and the non-RAG settings.

\begin{figure}
    \centering
    \includegraphics[width=0.75\linewidth]{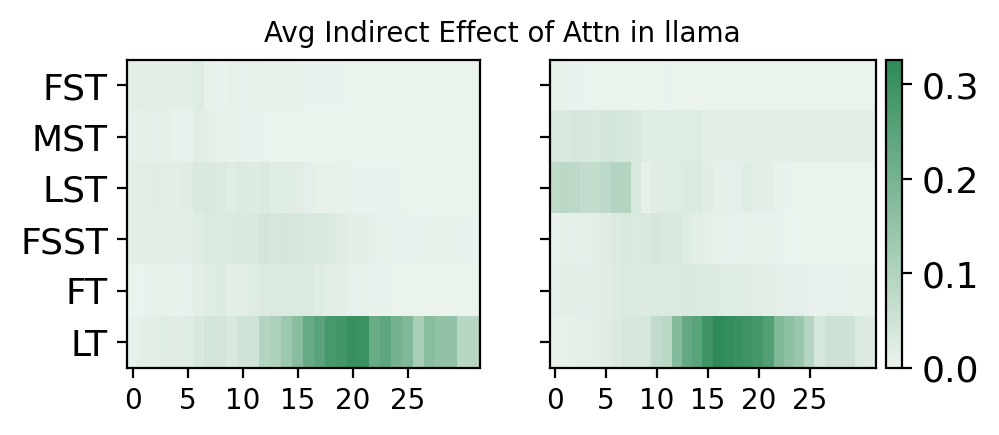}
    \caption{LLama-2 causal trace on Attention}
    \label{fig:llama_attention_causal_trace}
\end{figure}

\begin{figure}
    \centering
    \includegraphics[width=0.75\linewidth]{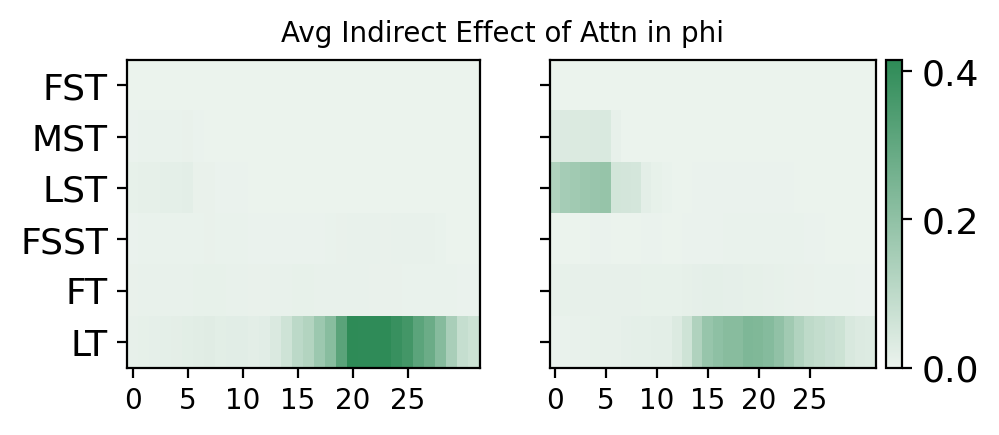}
    \caption{Phi-2 causal trace on Attention}
    \label{fig:phi_attention_causal_trace}
\end{figure}

In addition to causal tracing on MLPs, we also perform causal tracing on the attention modules, which we present in this section in \ref{fig:llama_attention_causal_trace} and \ref{fig:phi_attention_causal_trace}

We observe fairly similar traces for attention in the RAG vs non-RAG settings. The last token is crucial in both settings, thus effectively establishing that all information required for the task is written to the last token's residual stream, with the source being subject in the non-RAG case, and the source being the attribute token in the RAG setting. 

To apply noise to the token embeddings, we use the automatic spherical gaussian noise, the default setting used in \cite{meng2022locating}. The noise is sampled from a gaussian distribution of mean $0$ and standard deviation $\nu$ where $\nu = 3\sigma$, where $\sigma$ is the standard deviation of a sample of token embeddings.

\end{document}